\documentclass[conference]{IEEEtran}
\IEEEoverridecommandlockouts
\usepackage{cite}
\usepackage{amsmath,amssymb,amsfonts}
\usepackage{algorithmic}
\usepackage{graphicx}
\usepackage{textcomp}
\usepackage{xcolor}
\usepackage{subcaption}
\usepackage[a4paper, total={184mm,239mm}]{geometry}
\usepackage[raggedrightboxes]{ragged2e}
\def\BibTeX{{\rm B\kern-.05em{\sc i\kern-.025em b}\kern-.08em
    T\kern-.1667em\lower.7ex\hbox{E}\kern-.125emX}}

\begin{document}
\captionsetup{font=footnotesize}

\title{Compressing VAE-Based Out-of-Distribution Detectors for Embedded Deployment
\thanks{This research is part of the programme DesCartes and is supported by the National Research Foundation, Prime Minister’s Office, Singapore under its Campus for Research Excellence and Technological Enterprise (CREATE) programme.  This research was funded in part by MoE, Singapore, Tier-2 grant number MOE2019-T2-2-040.}
}

\author{
\IEEEauthorblockN{Aditya Bansal,\textsuperscript{1} Michael Yuhas,\textsuperscript{1,2} Arvind Easwaran\textsuperscript{1}}
\IEEEauthorblockA{
  \textit{\textsuperscript{1}College of Computing and Data Science} \\
  \textit{\textsuperscript{2}Energy Research Institute @ NTU, Interdisciplinary Graduate Program} \\
  \textit{Nanyang Technological University}, Singapore \\
  aditya018@e.ntu.edu.sg, michaelj004@e.ntu.edu.sg, arvinde@ntu.edu.sg}
}

\maketitle

\begin{abstract}
Out-of-distribution (OOD) detectors can act as safety monitors in embedded cyber-physical systems by identifying samples outside a machine learning model's training distribution to prevent potentially unsafe actions.  However, OOD detectors are often implemented using deep neural networks, which makes it difficult to meet real-time deadlines on embedded systems with memory and power constraints.  We consider the class of variational autoencoder (VAE) based OOD detectors where OOD detection is performed in latent space, and apply quantization, pruning, and knowledge distillation.  These techniques have been explored for other deep models, but no work has considered their combined effect on latent space OOD detection.  While these techniques increase the VAE's test loss, this does not correspond to a proportional decrease in OOD detection performance and we leverage this to develop lean OOD detectors capable of real-time inference on embedded CPUs and GPUs.  We propose a design methodology that combines all three compression techniques and yields a significant decrease in memory and execution time while maintaining AUROC for a given OOD detector.  We demonstrate this methodology with two existing OOD detectors on a Jetson Nano and reduce GPU and CPU inference time by 20\% and 28\% respectively while keeping AUROC within 5\% of the baseline.
\end{abstract}

\section{Introduction}
Deep learning models are usually trained with a closed-world assumption -- the data on which the model is trained is entirely representative of real-world data. However, when the models are deployed in real-world settings, the test data distribution might be drastically different from the training set, resulting in poor performance. Although this loss in accuracy might be tolerable for non-critical applications, for mission-critical applications like autonomous driving, such accuracy drops cannot be tolerated.  Hence, to ensure safety in machine learning models, it is necessary to detect out-of-distribution (OOD) samples at test time and relay this information to a high level controller, which can put the system into a safe state or hand control back to a human operator if possible. 

One of the key applications of OOD detection is in embedded systems where a deep neural network (DNN) has to be deployed to a resource constrained device while still being capable of making online inferences. Thus, it is critical to ensure that such models are efficient, with low inference times and storage requirements, along with high classification accuracy. Consider a mobile robot navigating an environment using a DNN.  If most of the training data was gathered during normal brightness conditions, performance may degrade in dark or extremely bright environments.  An OOD detector designed to protect this system needs to meet a minimum accuracy requirement, not exceed a maximum execution time, and achieve minimal memory usage so as not to interfere with other mission-critical tasks.  While many OOD detection architectures have been explored in previous literature~\cite{ruff2021unifying}, the variational autoencoder (VAE) based OOD detection architecture has received interest in the context of embedded systems due to its ability to learn interpretable representations in latent space~\cite{ramakrishna2022efficient}.

\begin{figure}
\includegraphics[width=0.49\textwidth]{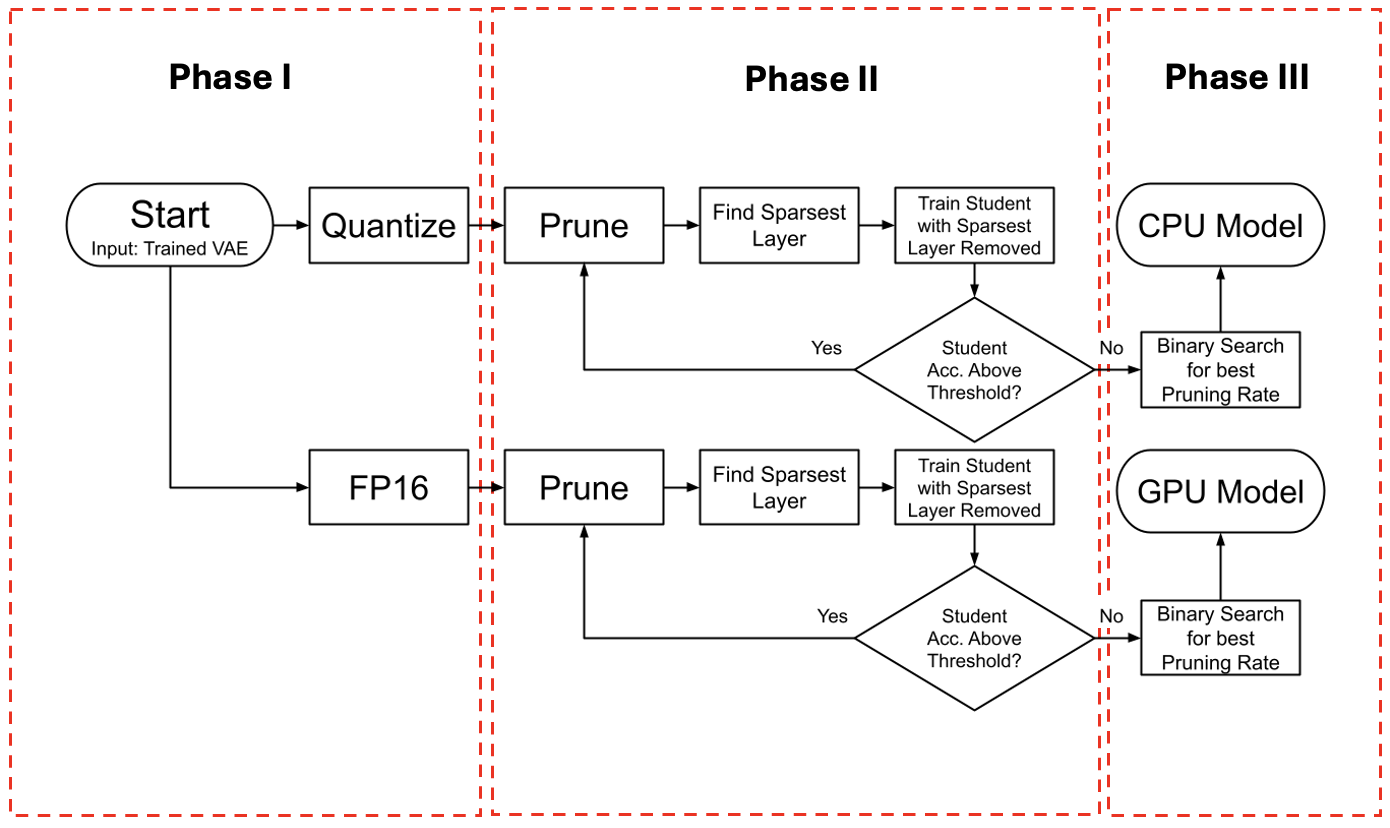}
    \centering
    \caption{Our design methodology for compressing a VAE-based OOD detector using pruning informed knowledge distillation and quantization.}
    \label{fig:framework}
\vspace{-5mm}
\end{figure}

Previous works have explored knowledge distillation~\cite{wu2022multi}, pruning~\cite{koda2023pros}, and quantization~\cite{yuhas2022design} for OOD detection, but none have focused on the combination of all three techniques.  VAE-based OOD detectors present an additional challenge as there is no direct relationship between loss and detection performance, unlike other models where compression was explored~\cite{wu2022multi, koda2023pros}. We show that these compression techniques applied to a VAE-based OOD detector can yield significant reductions in inference time and memory footprint without loss of accuracy for the OOD detection task. This is particularly useful for memory-intensive models in embedded systems \cite{chen2020deep}. Moreover, we observe that low training loss in the VAE does not necessarily translate to high OOD detection accuracy, and vice-versa.  This implies that the VAE-based OOD detector can maintain some level of detection performance, even when its loss is degraded by compression.

Based on our experiments, we propose the design methodology shown in Fig.~\ref{fig:framework} to compress an existing VAE-based OOD detector for deployment in an embedded system while maintaining acceptable classification accuracy.  Our framework relies on three stages: 1) quantization, 2) pruning informed knowledge distillation, and 3) final pruning.  As many modern SoCs are equipped with both CPU and GPU cores, our methodology generates two models: a floating point version tailored to GPU inference and a quantized version tailored to CPU inference.  We demonstrate the methodology on two VAE-based OOD detectors: $\beta$-VAE, which learns an interpretable representation of latent space~\cite{ramakrishna2022efficient} and optical flow (OF), which learns two latent representations for horizontal and vertical flows~\cite{feng2021improving}.  With the $\beta$-VAE model we reduce execution time by 38\% while AUROC only falls 1\% below baseline.  With the OF model, we reduce execution time by 20\% and 28\% on GPU and CPU respectively while maintaining AUROC within 5\% of baseline.

\section{Related Work}

Quantization has proven successful in both training~\cite{banner2018scalable} and neural network inference~\cite{gholami2021survey}. Quantization consists of techniques for reducing the precision of weights and activations in a neural network. By storing tensors at a lower bit-width than floating-point precision, drastic execution time improvements can be made in model inference and the storage space requirements of such models. Broadly, quantization can be categorized into three different algorithms: dynamic quantization, post-training static quantization, and quantization aware training (QAT)~\cite{nagel2021white}. The algorithms primarily differ in when and how the quantization parameters are determined. \textit{Dynamic quantization} computes quantized weights offline, but activations dynamically during inference time, \textit{quantization-aware training} trains a model with quantized parameters, and \textit{post-training static quantization} performs an additional calibration step after training. Quantized networks have shown impressive performance on various benchmark datasets such as MNIST, CIFAR-10 and ImageNet~\cite{hubara2017quantized}.

Neural network pruning removes the redundant and inconsequential parameters from a neural network, by either removing individual parameters or removing groups of parameters such as entire filters or channels from a convolutional layer. Based on the importance of the neurons or weights in a model, the ones that contribute the least to the output are removed, until the desired compression is achieved. Several pruning algorithms have been proposed in the literature to improve neural network efficiency (\emph{e.g.},~\cite{zhang2019eager,ponnapalli1999formal, augasta2011novel}). 

Knowledge distillation aims to train a simpler and smaller `student model' from a larger complex model or an ensemble of models with minimal loss in accuracy \cite{hinton2015distilling}. As noted by \cite{gou2021knowledge}, distillation algorithms typically differ in the form of knowledge, the distillation algorithm, and the training technique.

Wu \textit{et al.}, considered knowledge distillation for OOD detectors~\cite{wu2022multi} and Koda \textit{et al.} considered pruning~\cite{koda2023pros}, but only in cases where the OOD detector was integrated with a larger classification DNN.  In these studies the classification accuracy of the DNN was considered, not the interpretability of the OOD detector, which is a driving factor behind VAE-based OOD detection.  Furthermore, these studies did not consider how to combine compression techniques and did not feature quantization.

\section{Background -- OOD Detection}
\label{subsection:ood_detection}

 Designing accurate OOD detectors has been explored in previous literature. In order to deal with the high dimensionality of image data, OOD detectors have commonly been implemented as DNNs.  For example, in~\cite{linmans2020efficient}, multi-head CNNs were used to quantify uncertainty in a model's prediction, allowing an image to be classified as in-distribution (ID) or OOD, while in~\cite{valiuddin2022efficient} generative models with normalizing flows were used to perform OOD detection given some domain knowledge of ID characteristics. Another common method is to use a VAE, which learns a latent distribution of input samples and attempts to reconstruct them from this representation.  Reconstruction based methods~\cite{an2015variational} use this information bottleneck to ensure OOD samples are reconstructed poorly and can be identified easily during test.  Recently, detection in the latent representation space has become popular as it only requires executing the VAE encoder during test and yields better detection results~\cite{vasilev2020q}.

\begin{figure}
     \centering
     \begin{subfigure}{0.48\textwidth}
         \centering
         \includegraphics[width=1\textwidth]{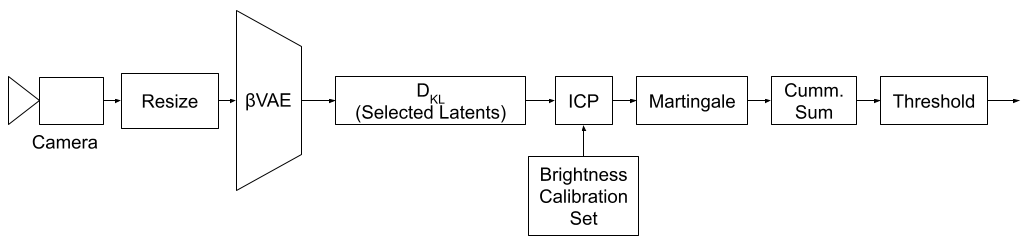}
         \caption{$\beta$-VAE OOD detector trained to detect shifts in brightness.}
         \label{fig:bvae_bd}
     \end{subfigure}
     \begin{subfigure}{0.48\textwidth}
         \centering
         \includegraphics[width=1\textwidth]{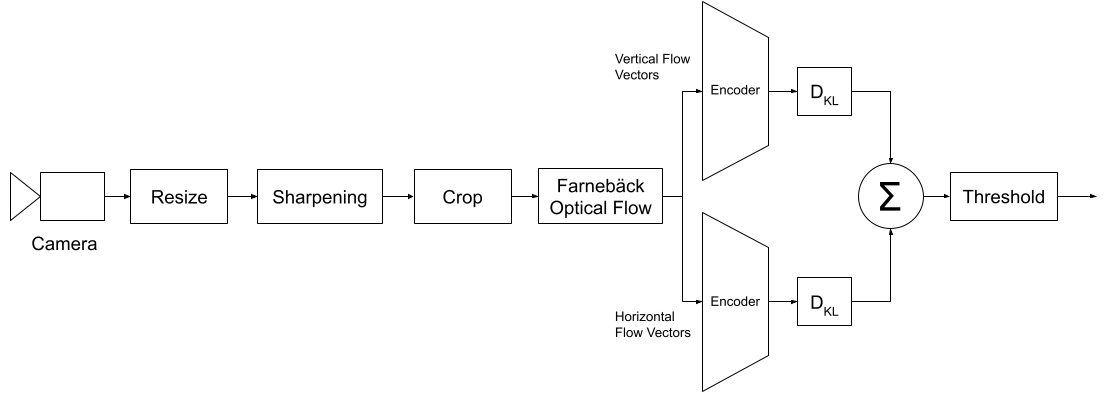}
         \caption{Optical flow OOD detector operating on horizontal and vertical flow vectors.}
         \label{fig:optflow_bd}
     \end{subfigure}
     \caption{Block diagrams of the OOD detectors considered in our case studies.}
     \label{fig:bd}
     \vspace{-5mm}
\end{figure}

The $\beta$-VAE OOD detector shown in Fig.~\ref{fig:bvae_bd} is one such latent space detection methodology~\cite{ramakrishna2022efficient}, and allows reasoning about OOD samples based on their latent distributions.  ID training data is divided into partitions based on generative factors, \emph{e.g.}, precipitation or ambient brightness. All partitions are used to train a VAE using a modified loss function that combines reconstruction loss with a Kullback-Leibler (KL) divergence term multiplied by factor $\beta$, which enforces a normal distribution of samples in the latent space~\cite{higgins2016beta}. After training, a set of calibration data containing samples from each data partition is fed into the network and the latent variables that show the most variance with respect to a particular generative factor are selected as reasoners for that factor.  At inference, the KL divergence between a latent reasoner and the standard normal distribution is fed into an inductive conformal prediction (ICP) algorithm~\cite{laxhammar2015inductive}, which assigns a confidence level as to whether a sample belongs to the same distribution as the calibration set. A martingale~\cite{fedorova2012plug} and cumulative sum are then used to smooth these confidence scores over time as environmental conditions like precipitation and brightness are unlikely to change instantaneously between frames.

Another example of latent space OOD detection is the optical flow OOD detector shown in Fig.~\ref{fig:optflow_bd}~\cite{feng2021improving}.  This OOD detector is built on the notion that certain motions in a vehicle's environment have been encountered during training and other OOD motions indicate the presence of hazardous conditions. First, the Farneb{\"a}ck optical flow is calculated between two sequential input frames, which generates a vector field of flows with horizontal and vertical components~\cite{farneback2003two}. The vector fields for six successive frames are concatenated and used as input to two VAEs: one for the vertical flow components and another for the horizontal ones. During training these VAEs use the evidence lower bound (ELBO) loss function to reconstruct the original flow vectors and enforce normality in the the latent space~\cite{higgins2016beta}. At test time, KL divergence is used to measure how far a sample deviates from the expected distribution. Because a sample can be OOD with respect to horizontal or vertical flows, the KL divergence for both the encoders are summed and compared against a specified threshold.

\section{Design Methodology}

VAE-based OOD detection methods require inferencing an encoder network for each sample at run time.  This becomes a challenge in systems with hard deadlines where these models must process large input images.  Previous works have analyzed reducing the dimensionality of these models' inputs~\cite{yuhas2022design}, however, this may discard  potentially useful information in an input sample.  Our methodology formulates the problem differently: given an input with fixed dimensionality and data format, how do we design an OOD detector that minimizes execution time and memory usage while maintaining accuracy above a given threshold.  The search space for this design problem is prohibitively large (quantization level, pruning amount, and student architecture for knowledge distillation).  To address this, our methodology starts with a greedy search through potential student architectures, testing those most likely to lead to a reduction in execution time while maintaining accuracy first.  Afterward, a binary search through pruning levels is used to select the minimum memory configuration for a given architecture that satisfies our accuracy constraints.  All these steps require access to a limited number of OOD samples for cross-validation purposes, even though the OOD detectors are trained using only ID samples.  The components of our methodology from Fig.~\ref{fig:framework} are explained in detail below.

\textbf{Target Aware Model Generation} - Different embedded hardware have different requirements for the minimum quantization level and precision that can be used efficiently at run time.  For example, an embedded GPU may only support floating point operations, while a CPU may require quantized neural operations in order to achieve a reasonable execution time.  Furthermore, previous works targeted at embedded deployment require a model to have two versions: one for embedded CPUs and another for embedded GPUs depending on resource availability at run time~\cite{ling2022blastnet}.  Our methodology achieves this by separately designing a 16-bit floating point (fp16) model for an embedded GPU and a quantized 8-bit (qint8) model for CPU.  The design process is separate for each model as different precision levels may tolerate different amounts of pruning and knowledge distillation before violating accuracy constraints.

\textbf{Pruning Aware Knowledge Distillation} - Training a student network that requires fewer operations than its teacher model has the largest potential to reduce execution time.  While the impacts of quantization and precision reduction are hardware dependent, gains from knowledge distillation are universally applicable.  However, the search space of possible student models is prohibitively large, so we propose pruning aware knowledge distillation to define the subset of models to be explored and greedily searched within this space.  First, we propose that all student models should be based on the teacher architecture and that execution time decreases should come from the removal of layers.  First, 50\% of the weights and biases with the lowest L2 mean in the teacher model are pruned and set to zero.  Then, the layer with the highest percentage of zero weights is removed and the resulting model becomes the student.  If a layer's bias vector has the highest percentage of zeros, then the bias vector is removed.  When a layer is removed, the output of the preceding layer must match the shape of the next layer in the original model.  To accomplish this for convolutional layers without causing an explosion in the number of learnable parameters, stride and dilation are increased while kernel size and number of input and output channels remain unchanged.  For linear layers, the number of outputs is simply set to the the number of inputs for the next layer.  We take a layer's sparsity after pruning as a heuristic for its contribution to the network's output; in this way we perform greedy search, removing layers we believe correspond the least to the model's output first, to quickly arrive at the architecture that best satisfies the design requirements.

\textbf{Obtaining the Sparsest Model} - Pruning aware knowledge distillation stops when the pruned model no longer satisfies the accuracy constraint.  We take the last student model from the previous step that satisfied all constraints and perform a binary search across sparsity (from $0-100\%$) to find a model with the greatest possible sparsity that still satisfies all constraints.  First, the model is pruned to the desired sparsity,~$s$.  If the accuracy is greater than the constraint, the model is pruned again such that its new sparsity is $(s + (100-s)/2)\%$, otherwise the unpruned model is pruned to $(s/2)\%$ sparsity and the process repeats.  The assumption is that sparser models will have worse accuracy than less-sparse models, but there will exist some sparsity level at which the student model will still satisfy the accuracy constraint, but benefit from the the storage size reduction resulting from pruning.

\section{Case Study: $\beta$-VAE OOD Detector}

\begin{figure}
     \centering
     \begin{subfigure}{0.4\textwidth}
         \centering
         \includegraphics[width=1\textwidth]{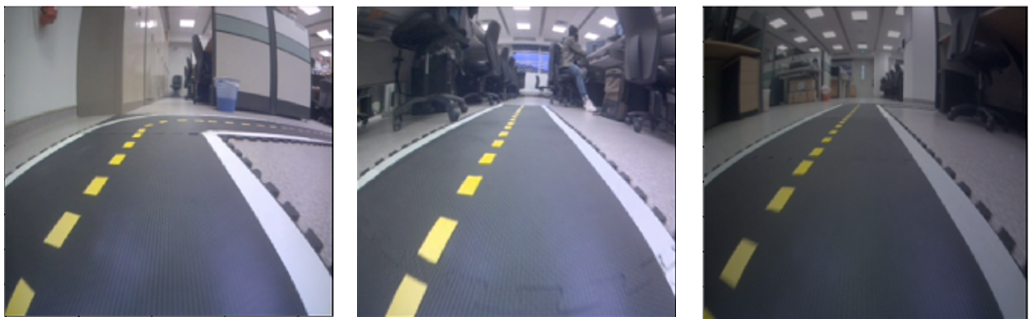}
         \caption{Original input images}
         \label{fig:orig_imag}
     \end{subfigure}
     \begin{subfigure}{0.4\textwidth}
         \centering
         \includegraphics[width=1\textwidth]{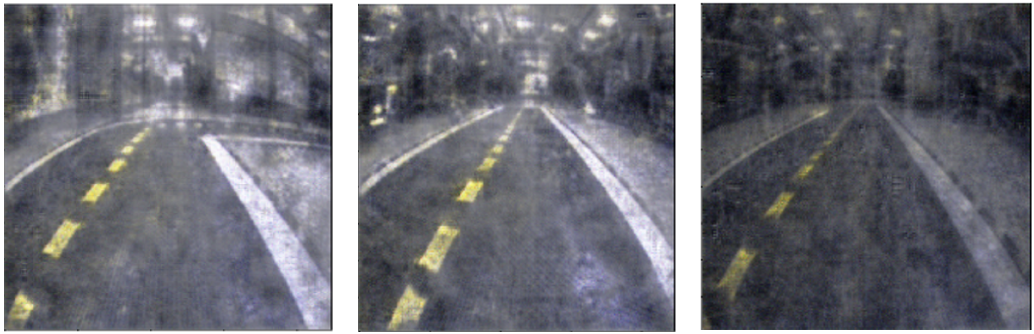}
         \caption{Reconstructed Images}
         \label{fig:recon_img}
     \end{subfigure}
     \caption{Sample images reconstructed using the $\beta$-VAE model}
     \label{fig:bd}
     \vspace{-5mm}
\end{figure}

A miniature robotic platform, called Duckietown \cite{paull2017duckietown}, was used as a test-bed to simulate autonomous driving. To validate the performance of the VAE-based OOD Detector, a test vehicle, called a Duckiebot (Jetson Nano quad-core ARM Cortex-A57 64-bit CPU @ 1.42 GHz with 2 GB RAM), was deployed in the Duckietown environment to collect data in different brightness conditions. The data was collected on four different tracks (three indoor and one outdoor track), under three ambient lighting conditions -- low, medium, and high -- as illustrated in Figure \ref{fig:orig_imag}. Low and medium brightness levels were considered as in-distribution (ID), and high brightness was considered as OOD data. A calibration set was generated from the same distribution as that of ID data. This set consisted of rapidly fluctuating brightness levels between low and medium ID partitions to allow the model to identify the latent variables corresponding to brightness and rank them based on the amount of variance. Similar to \cite{ramakrishna2022efficient}, a 2:1 training to calibration ratio was used during calibration.  

A beta variational auto-encoder ($\beta$-VAE) architecture, as proposed by \cite{ramakrishna2022efficient}, was trained for 200 epochs using ELBO loss and optimised using the Adam optimizer. The encoder consisted of four convolution blocks each having a 2D convolution layer followed by 2D batch normalisation, leaky ReLU activation function, and finally a 2D max pooling layer. At the end, four fully-connected layers were used with leaky ReLU activation functions. The decoder block, typically an inverse of the encoder block, consisted of four fully connected linear layers with leaky ReLU activation functions, with an increasing number of neurons in each subsequent layer. The fully-connected layers were followed by four deconvolution blocks, each comprising of max unpooling layer, transpose convolution, batch normalization, and leaky ReLU activation function. The number of latent dimensions (30) and $\beta$ value (1.4) were determined using Bayesian optimization to maximize the VAE's mutual information gain as recommended by~\cite{ramakrishna2022efficient}.

As discussed in Section \ref{subsection:ood_detection}, the $\beta$-VAE model is first trained on ID data gathered in different lighting conditions, and the weights of the decoder part are discarded and the encoder-only model is used for OOD detection. Figure \ref{fig:recon_img} illustrates sample images reconstructed using the baseline $\beta$-VAE model. The baseline model achieved a true positive rate of $0.955$ and a false positive rate of $0.250$. The area under the ROC curve (AUROC) for the optimum decay values was $0.852$.

\subsection{Compressing the $\beta$-VAE OOD Dectector}

The compression techniques of quantization, pruning, and knowledge distillation were independently explored on the $\beta$-VAE OOD detector, and the results are presented in figure \ref{fig:all_summary}.

\noindent{\textbf{{Quantization}} - Table \ref{table:quant_perf} summarises the memory usage and execution times of three different quantization techniques. All quantization methods result in a memory improvement of about 3.7 times due to the conversion of weight precision from \textit{float32} to \textit{int8} in the quantized form. Dynamic quantization incurs the longest execution time, which could be attributed to the additional overhead of computing the scaling factors during runtime.  

As observed in Figure \ref{fig:all_summary}, dynamic quantization results in the least total loss, since the quantization takes place during runtime (`on-the-fly'), and hence it can effectively compute the quantization scaling factor for each instance. Further, quantization aware training results in the highest AUROC value, among the three techniques.

\begin{table}[]
\caption{Memory consumption and execution time of the model after Dynamic Quantization, Static Quantization and Quantization Aware Training (QAT)}
\label{table:quant_perf}
\begin{tabular}{|p{0.9in}|p{0.6in}|p{0.7in}|p{0.6in}|}
\hline
\textbf{Model} & \textbf{Model Size (MB)} & \textbf{Forward pass size (MB)} & \textbf{Execution Time (ms)} \\ \hline
Dynamic   Quantization        & 18.34   & 90.56 & \(90.39 \pm 4.05\)                        \\ \hline
Static   Quantization         & 17.75   & 40.47 & \(56.14 \pm 1.56\)                        \\ \hline
QAT & 17.76   & 40.48 & \(57.48 \pm 2.27\)                         \\ \hline
\end{tabular}
\vspace{-5mm}
\end{table}

\noindent{\textbf{Pruning}} - A global unstructured pruning was performed on the $\beta$-VAE model, where individual nodes with the lowest L2 mean weights were pruned away. Figure \ref{fig:pruning_plot} illustrates the reconstruction loss of the network across different sparsity levels. It can be observed that the total loss does not deteriorate until about 60\% of the nodes are pruned away. However, a significant drop in the AUROC is observed after pruning about 50\%. Such a drop could be attributed to the unstructured nature of the pruning technique used, where any neuron in the entire architecture could be pruned. As discussed in section \ref{section:case_study_2} a structured layer-wise pruning strategy has been opted to mitigate such a drastic drop in AUROC.

\begin{figure}[h]
\includegraphics[scale=0.23]{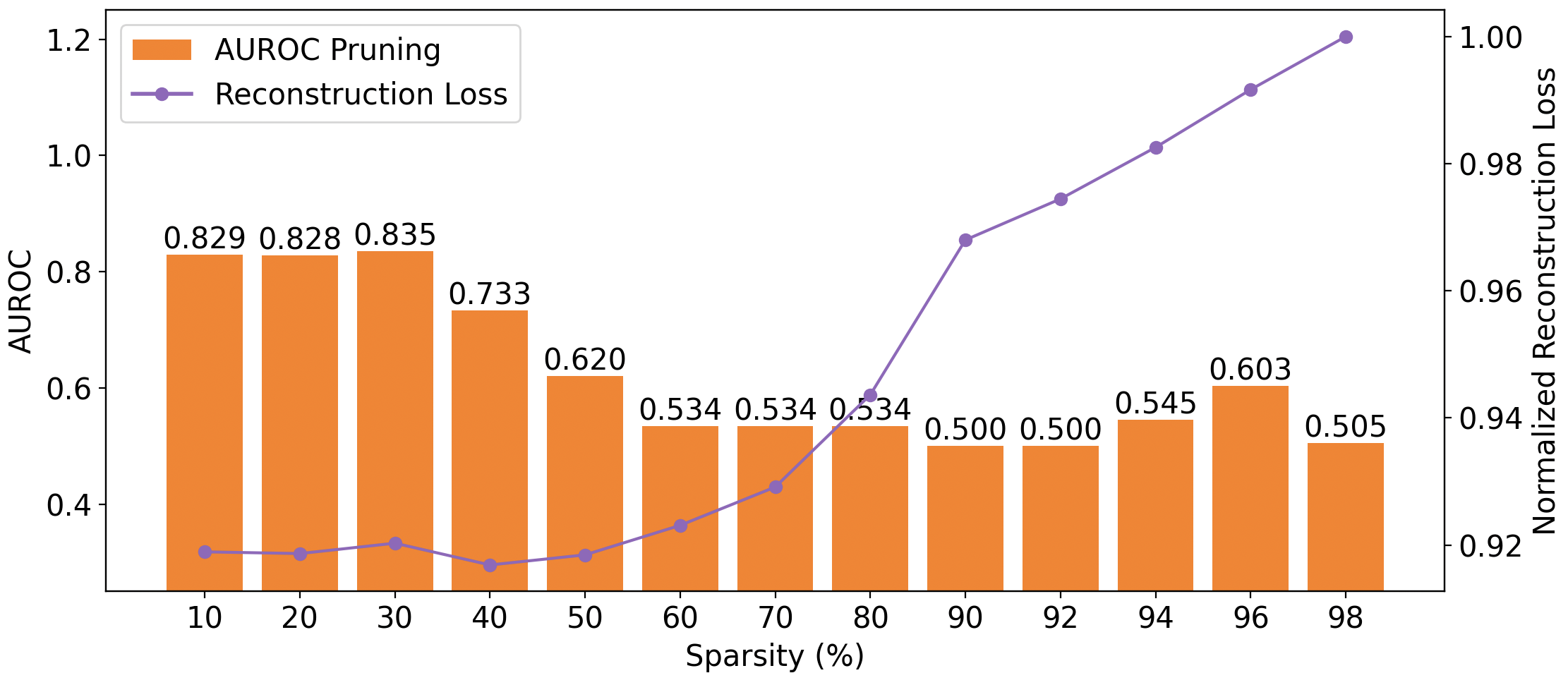}
    \centering
    \caption{AUROC and total reconstruction loss for the $\beta$-VAE detector at different sparsity levels using pruning}
    \label{fig:pruning_plot}
\end{figure}

\noindent{\textbf{Knowledge Distillation}} - Knowledge distillation was performed on a student architecture with one of the fully-connected layers removed, resulting in a reduction of more than 50\% in the number of parameters in the network. As demonstrated in table \ref{table:kd_mem}, the resultant student architecture had approximately 7 million parameters as opposed to the initial 17 million parameters. A second student architecture was tested with the removal of the last convolution layer in the encoder and the first deconvolution layer in the decoder part of the network. This resulted in a significant drop in the number of parameters, with just 400 thousand parameters in the entire network due to the reduced output shape in the absence of a convolution layer, which cascades to all the following fully connected layers. The first student architecture led to a relatively better AUROC performance, as compared to the second architecture, which reiterates the importance of the convolutional layer in encoding crucial information necessary for OOD detection.

The ablation studies shed light onto the influence of individual compression techniques on the overall system design. The experiments  reveal that low training loss does not necessarily translate to a high OOD accuracy. Comparison between dynamic quantization and quantization aware training reveal that although dynamic quantization leads to a lower reconstruction loss, the AUROC achieved by quantization aware training is higher than that of dynamic quantization. Overall, quantization results in the best OOD performance among all compression techniques. Interestingly, quantization aware training  surpassed the baseline AUROC, which could be attributed to the higher generalizability of quantized networks. Moreover, although the $\beta$-VAE network in our model was trained with a combination of reconstruction loss and KL loss, only KL divergence in the latent space is used at test time, which may have contributed to these compression techniques maintaining relatively high OOD detection accuracy (and sometimes leading to an improvement), despite causing an overall decrease in reconstruction loss.

\begin{table}[]
\caption{Memory footprint of the $\beta$-VAE model after knowledge distillation}
\label{table:kd_mem}
\centering
\begin{tabular}{|l|l|l|}
\hline
\textbf{Model}                    & \textbf{Number of Parameters} & \textbf{Model Size (MB)} \\ \hline
Base   Model             & $\sim$17 M & 67.42 \\ \hline
Student   Architecture 1 & $\sim$7 M  & 26.71 \\ \hline
Student   Architecture 2 & $\sim$400K & 1.61  \\ \hline
\end{tabular}

\end{table}

\begin{figure}
\includegraphics[width=0.49\textwidth]{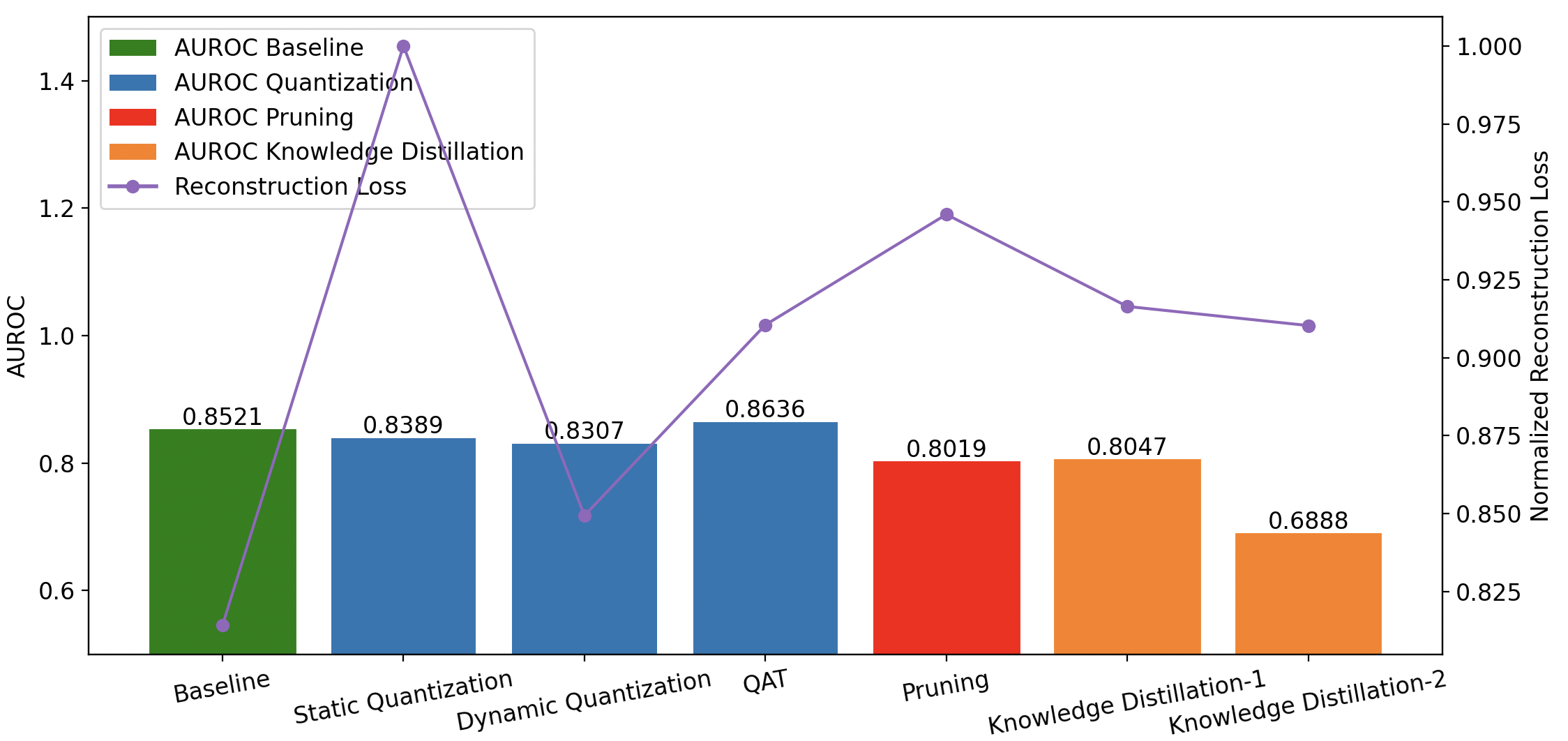}
    \centering
    \caption{Reconstruction loss and AUROC of the $\beta$-VAE detector across different compression techniques}
    \label{fig:all_summary}
    \vspace{-5mm}
\end{figure}

\section{Case Study: Optical Flow OOD Detector}
\label{section:case_study_2}
We tested our strategy on the optical flow OOD detector~\cite{feng2021improving}.  Using the same dataset as the previous case study, we artificially added rain and snow to 50\% of the images in the test set using the same method as~\cite{yuhas2022design}, such that samples with no precipitation were considered ID, while samples with precipitation were considered OOD. Both encoder networks (Fig.~\ref{fig:optflow_bd}) were trained with the same initial architecture referred to as the baseline.  This consists of 4 convolutional layers with stride 3, kernel size 5, and depths 32/64/128/256.  Each convolutional layer is followed by a batch norm and ReLU activation function.  Finally, a fully connected layer with linear activation leads to a latent space with 12 dimensions.  Both horizontal and vertical flow vectors were sized $224\times224$ with 6 flows concatenated to form one input.  During training, the decoder was the mirror image of the encoder.  Each network was trained for 100 epochs using the Adam optimizer with learning rate capped at $1\times10^{-5}$.

First, we performed our pruning aware knowledge distillation with the results shown in Fig.~\ref{fig:opt_flow_kd}.  Configuration (1) corresponds to the baseline model, in configuration (2) we retain convolutional layers with depths 32/64/256 and the linear layer, in configuration (3) we have convolutional layer depths 32/64/256 with no linear layer, in configuration (4) we have convolution depths 32/64, and in configuration (5) we only have a single convolution layer that takes input space to 12 output latent dimensions.  As expected, execution time (ET) decreases as layers are systematically removed, however, AUROC does not sharply fall off for the fp16 model until only one layer is left.  The qint8 model has the largest improvement in execution time for removing layers, but its AUROC also falls off much faster.  Surprisingly, for very few layers (configurations 4 and 5) it ends up performing better than some configurations with more free parameters.  We hypothesize that the initial effective capacity of the architecture was much less than its representational capacity, and for the configurations with fewer layers, even though they have less representational capacity, their effective capacity increases due to the architecture.  We also show that while removing layers greatly impacts the VAE's loss, good OOD detection performance is maintained because the mean KL loss on OOD samples remains above that of ID samples; as long as loss increases equally for all samples, performance is preserved.

\begin{figure}
\includegraphics[width=0.49\textwidth]{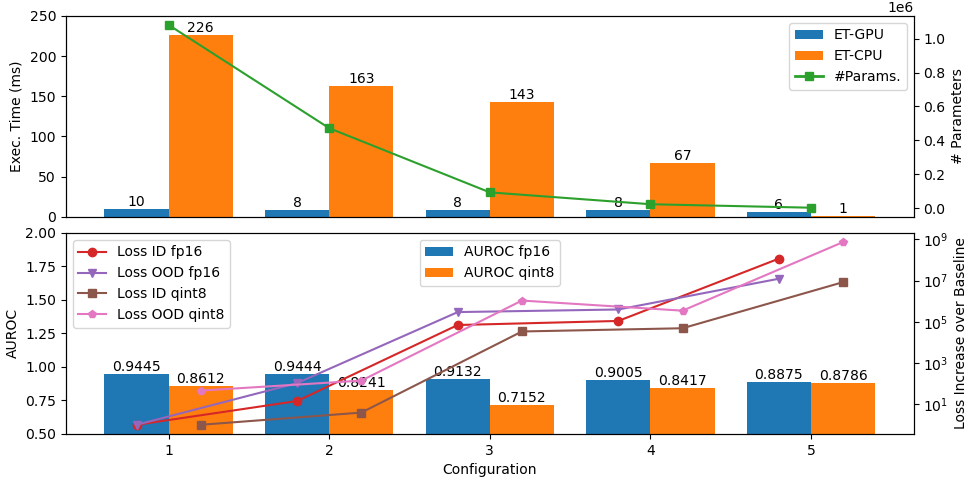}
    \centering
    \caption{AUROC, number of parameters, execution time (ET), and mean KL divergence loss for the optical flow OOD detector under five different knowledge distillation configurations: (1) baseline; (2) layer 4 removed; (3) layers 4 and 5 removed; (4) layers 3, 4, and 5 removed; (5) layers 2, 3, 4, and 5 removed.}
    \label{fig:opt_flow_kd}
    \vspace{-5mm}
\end{figure}

Since AUROC dips below 0.9 after configuration (4), we choose this model to analyze the effects of further pruning on performance.  Fig.~\ref{fig:prune_search_of} shows AUROC and corresponding loss on ID and OOD samples across varying sparsity levels for the fp16 and qint8 models.  These sparsity levels are calculated across all layers in the model.  We observe that increasing sparsity leads to an increase in loss, but because mean loss increases proportionally for ID and OOD samples, performance is preserved.  However, for very sparse models (e.g. 90\%) AUROC and loss both drop sharply.  This collapse in loss is explained by the fact that nearly all the weights are now 0, making it less likely that any given sample will lead to a high mean and variance, which would increase KL divergence in latent space.  Because the KL loss collapses for all samples regardless of ID or OOD, AUROC is also severely impacted.

\begin{figure}
\includegraphics[width=0.49\textwidth]{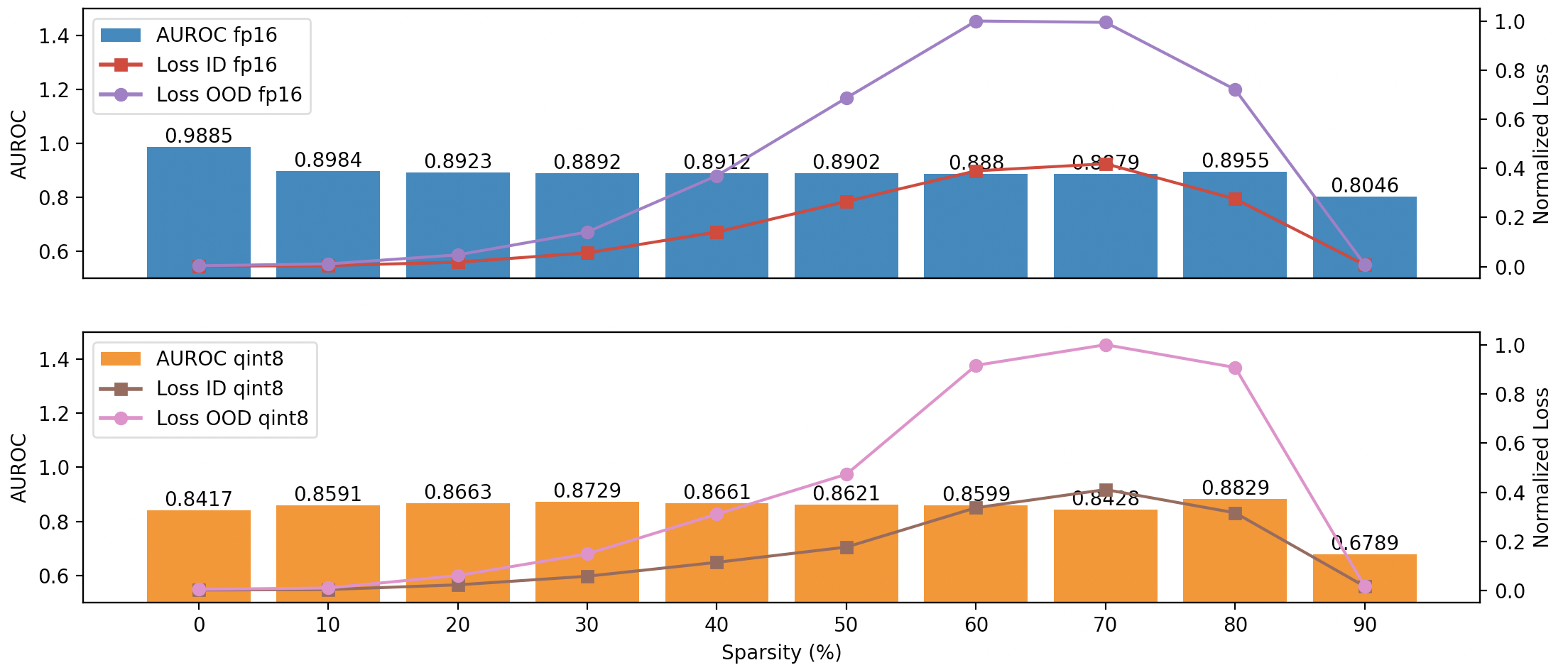}
    \centering
    \caption{AUROC and corresponding test loss for quantized and fp16 optical flow models under increasing sparsity.}
    \label{fig:prune_search_of}
    \vspace{-5mm}
\end{figure}

\section{Conclusion}

We explored different neural network compression techniques on $\beta$-VAE and optical flow OOD detectors using a mobile robot powered by a Jetson Nano. Based on our analysis of results for quantization, knowledge distillation, and pruning, we proposed a design strategy to find the model with the best execution time and memory usage while maintaining some accuracy metric for a given VAE-based OOD detector.  We successfully demonstrated this methodology on an optical flow OOD detector and showed that our methodology's ability to aggressively prune and compress a model is due to the unique attributes of VAE-based OOD detection.  

Despite our methodology's good performance, it requires access to OOD samples at design time to act as a cross-validation set.  In our case study, we  assume OOD samples arise from a particular generating distribution, but this may not be the case in general.  Furthermore, it only guides the search for a faster architecture, but does not guarantee the optimum result.  Nevertheless, we believe having a design methodology that combines quantization, knowledge distillation, and pruning allows engineers to exploit the combined powers of these techniques instead of considering them individually.

\bibliographystyle{IEEETran}
\bibliography{refs}

\end{document}